\documentclass[twoside, conference]{IEEEtran}

\IEEEoverridecommandlockouts                              % This command is only needed if 
\title{\LARGE \bf
Accurately Predicting Probabilities of Safety-Critical Rare Events for Intelligent Systems
}

\author{Ruoxuan Bai$^{1}$, Jingxuan Yang$^{2}$, Weiduo Gong$^{3}$, Yi Zhang$^{4}$,~\IEEEmembership{Senior Member,~IEEE, } Qiujing Lu$^{5}$, \\
Shuo Feng$^{6\ast}$, ~\IEEEmembership{Member,~IEEE}%
	\thanks{*This work is supported by Beijing Natural Science Foundation under Grant 4244092 and Beijing Nova Program under Grant 20230484259. \textit{(Corresponding author: Shuo Feng)}}% 
	\thanks{$^{1,2,3,5}$Ruoxuan Bai, Jingxuan Yang, Weiduo Gong and Qiujing Lu are with the Department of Automation, Tsinghua University, Beijing 100084, China {\tt\small ({brx22, yangjx20,gwd21,qiujinglu}@mails.tsinghua.edu.cn)}.}%
	\thanks{$^{4,6}$Yi Zhang, Shuo Feng are with the Department of Automation, Beijing National Research Center for Information Science and Technology (BNRist), Tsinghua University, Beijing 100084, China {\tt\small (zhyi, fshuo@tsinghua.edu.cn)}.}%
}

\usepackage{amsmath}
\usepackage{amssymb}
\usepackage{graphicx}
\usepackage{subfigure}
\usepackage{cite}
\usepackage{csquotes}
\usepackage{bm}
\begin{document}

\maketitle
\thispagestyle{empty}
\pagestyle{empty}

%%%%%%%%%%%%%%%%%%%%%%%%%%%%%%%%%%%%%%%%%%%%%%%%%%%%%%%%%%%%%%%%%%%%%%%%%%%%%%%%
\begin{abstract}
Intelligent systems are increasingly integral to our daily lives, yet rare safety-critical events present significant latent threats to their practical deployment. Addressing this challenge hinges on accurately predicting the probability of safety-critical events occurring within a given time step from the current state, a metric we define as \enquote{criticality}. The complexity of predicting criticality arises from the extreme data imbalance caused by rare events in high dimensional variables associated with the rare events, a challenge we refer to as the curse of rarity. Existing methods tend to be either overly conservative or prone to overlooking safety-critical events, thus struggling to achieve both high precision and recall rates, which severely limits their applicability. This study endeavors to develop a criticality prediction model that excels in both precision and recall rates for evaluating the criticality of safety-critical autonomous systems. We propose a multi-stage learning framework designed to progressively densify the dataset, mitigating the curse of rarity across stages. To validate our approach, we evaluate it in two cases: lunar lander and bipedal walker scenarios. The results demonstrate that our method surpasses traditional approaches, providing a more accurate and dependable assessment of criticality in intelligent systems.

\end{abstract}

%%%%%%%%%%%%%%%%%%%%%%%%%%%%%%%%%%%%%%%%%%%%%%%%%%%%%%%%%%%%%%%%%%%%%%%%%%%%%%%%

\begin{figure*}[!h]
	\centering
	\includegraphics[scale=0.33]{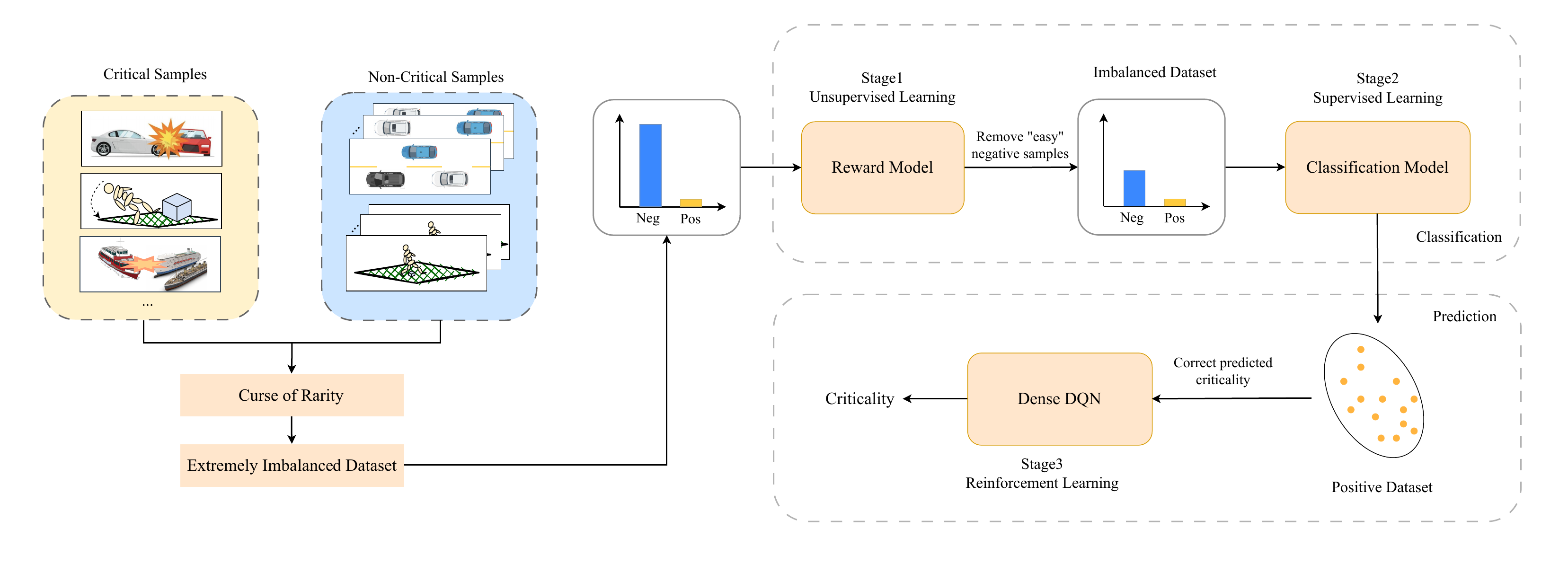}
	\label{1}
	\caption{Overview of the multi-stage learning framework. Our approach consists three stages. In the first stage, we remove those obvious non-critical samples and reduce imbalance ratio by unsupervised learning. Then in the second stage, we use labeled data to train a supervised classification model to further categorize those samples unable to be distinguished by unsupervised learning model. Lastly, in the third stage, we turn to improve the accuracy of predicted criticality other than continuously focusing on unclassified samples. Dense DQN method is developed to fine-tune last layers of classification model.}
\end{figure*}
\section{INTRODUCTION}
% safety critical system很重要
Intelligent systems are playing an increasingly significant role in our daily lives. Safety-critical autonomous systems are the intelligent systems whose failure will cause considerable losses of human life and property, such as autonomous vehicles, robotics, and intelligence medical diagnosis system\cite{cerrolaza2020multi,cummings2020regulating,guiochet2017safety}. Given the potential catastrophic consequences, it is imperative to guarantee these systems operate safely and effectively for practical deployment.

% 为什么高精准率高召回率这么重要
To address the aforementioned challenges, the key lies in precisely predicting the probabilities of safety-critical events associated with their failures within a given time step from current state, a metric we refer to as \enquote{criticality}. These events may span from a traffic accident involving an autonomous vehicle to a robot tipping over.  Previous criticality prediction methods can be broadly categorized into metrics-based methods and model-based methods.

The metrics-based criticality prediction methods rely on specific metrics designed to directly quantify risk, for instance the well-known Time To Collision (TTC) metric used in automated driving to assess the probability of collision\cite{westhofen2023criticality,feng2021parti,feng2021partii}. However, these metrics may not always yield precise results, and formulating appropriate metrics in complex systems can be a challenging task. In such scenarios, predictive models trained on data have emerged as a viable alternative. Although criticality prediction models have garnered significant attention in the field of disease diagnosis\cite{cao2023large,ratul2022early,jena2021risk}, achieving precise predictions remains  exceptionally challenging.

% 核心问题挑战
In this study, we aim to develop a criticality prediction model for intelligent systems that achieves both high precision and recall rates. A high recall rate is essential to reduce missed alarms, ensuring that all safety-critical events can be identified and timely measures can be taken to avert potential safety hazards. A high precision rate helps mitigate false alarms, preventing the systems from being overly conservative. 
The primary challenge arises from the rarity of safety-critical events in high-dimensional scenarios , often referred to as the \enquote{curse of rarity}\cite{liu2022curse}. This rarity results in an extremely imbalanced dataset, presenting significant obstacles in learning a well-performed model. 

% 现有方法简介
A long line of work have been proposed to overcome the imbalanced dataset issues. One popular approach is data re-sampling strategies, which often have difficulties in learning robust and generalizable features. The over-sampling method \cite{ando2017deep,pouyanfar2018dynamic,shen2016relay} can lead to overfitting on the minority class, while the under-sampling method\cite{buda2018systematic,lee2016plankton} may cause significant information loss in the majority class, leading to underfitting. Another approach is data re-weighting method \cite{byrd2019effect,cao2019learning,cui2019class,dong2018imbalanced,huang2016learning,khan2019striking}, where different weights are adaptively assigned to different classes or samples. Nonetheless, a potential problem is that the classifier may converge to a local optimal solution if the batches are dominated by majority class samples. These methods often yield a considerable number of false alarms, thereby resulting in a low precision rate.
 Recently, researchers have introduce decoupling method, meta-learning, transfer learning and contrastive learning\cite{kang2019decoupling,jamal2020rethinking,shu2019meta,liu2019large,yin2019feature,shi2023clip} to enhance imbalanced learning.
However, despite these efforts, existing approaches have struggled to obtain models with high precision and recall rate when confronted with extremely imbalanced testing dataset.

% 我们的方法
To effectively resolve such issues, we propose a multi-stage learning framework that gradually densifies the dataset and alleviates the degree of imbalance. 
As illustrated in Fig. 1, in the first stage, we employ unsupervised learning methods to filter out \enquote{easy} negative samples, the criticality of which is considered to be zero.
Subsequently, in the next stage, we turn to supervised learning and the enhanced bilateral-branch network (BBN)\cite{zhou2020bbn} is designed to perform more fine-grained classification task. 
Intuitively, samples difficult to classify in one feature space may become easier to separate in another space. By combining supervised and unsupervised methods, samples are projected into different state spaces for classification, enabling a coarse-to-fine identification of safety-critical events. 
Eventually, dense reinforcement learning method\cite{feng2023dense} is introduced to further improve the precision of predicted criticality. The effectiveness of dense reinforcement learning stems from the utilization of a more balanced gradient to fine-tune classification model in the second stage. 

% case study
To evaluate the effectiveness of our proposed method, we conducted experiments in two different scenarios and compared our approach with classical methods. The results demonstrate that our predictive model achieves high precision and recall rate, outperforming traditional approaches.
Our contributions can be summarized as follows:
\begin{itemize}
	\item To address the challenges posed by the curse of rarity, we develop a multi-stage learning framework to learn criticality prediction models with high precision and recall rate by progressively mitigating the imbalance ratio and enhancing accuracy.
\end{itemize}
\begin{itemize}
	\item Our method exhibits applicability to real-world scenarios featuring extremely imbalanced test datasets, with an imbalance ratio exceeding $10^4$.
\end{itemize}

The rest of this paper is organized as follows. The problem of criticality prediction is formulated in Section II. Our approach and algorithms are described in Section III. In Section IV, we analyze experiment results. Finally, Section V summarizes our proposed method and prospects for future work.

\section{PROMBLEM FORMULATION}

In this section, we formulate the problem of criticality prediction. If we denote a critical event as $A$ and the state of systems and environment as $\bm{X}$, the task of criticality prediction essentially involves estimating the conditional probability $\mathbb{P}(A|\bm{X})$. The criticality prediction we study in this work is in essence a rare-event probability estimation problem in a high-dimensional space due to the curse of dimensionality and the curse of rarity. The curse of dimensionality refers that state $\bm{X}$ is spatially and temporally complex, requiring numerous high-dimensional variables for accurate representation. The \enquote{curse of rarity}, which is the primary source of our challenges, refers to the infrequent occurrence of the critical event $A$.  This rarity implies that the majority of points within the variable space do not represent critical situations, leading to unreliable and noisy data.

Firstly, we formulate the problem of criticality prediction as a classification task. The objective is to train a classification model that can accurately determine whether a critical event $A$ will occur  given the current state $\bm{X}$ of the system. To this end, suppose we have a dataset $\mathcal{D}$ containing $N$ samples $ \{\bm{X}_k,\bm{Y}_k\}_{k=1}^N$.  Each sample $\bm{X}_k$ comprises the state of the intelligent system and its surrounding environment. The corresponding label $\bm{Y}_k$ are binary, taking values from $\{0,1\}$. These data is labeled by

\begin{equation}
	\bm{Y}_k = \mathbb{I}_A(\bm{X}_k), 
\end{equation}
where $Y_k=1$ if the critical event $A$ occurs within a certain time step given the state $X_k$ and $Y_k=0$ otherwise.

We define positive dataset $\mathcal{P}$ and negative dataset $\mathcal{N}$ as follows
\begin{equation}
	\mathcal{P} = \{(\bm{X},y) \in \mathcal{D} |y=1\},
\end{equation}
\begin{equation}
	\mathcal{N} = \{(\bm{X},y) \in \mathcal{D} |y=0\}, 
\end{equation}
In this study, the positive class refers to the minority class. For extremely imbalanced datasets, the number of samples belonging to the positive class ( $|\mathcal{P}|$) is significantly smaller than the number of samples belonging to the negative class ( $|\mathcal{N}|$). To quantify the degree of imbalance, the imbalance ratio (IR) is introduced, defined as 
\begin{equation}
	\mathrm{IR} = \frac{|\mathcal{N}|}{\mathcal{|P|}}.
\end{equation}

The imbalance ratio studied  in this work exceeds $10^{4}$,  a magnitude considerably higher than most existing studies. Such extreme class imbalance can cause the classifier to exhibit significant bias toward the negative class, resulting in a high false positive rate. The probability predicted by the classification model that a sample $\bm{X}$ belongs to positive class serves as an estimation of criticality $\mathbb{P}(A\mid \bm{X})$ . 

\begin{figure}[t]
	\centering
	\includegraphics[scale=0.5]{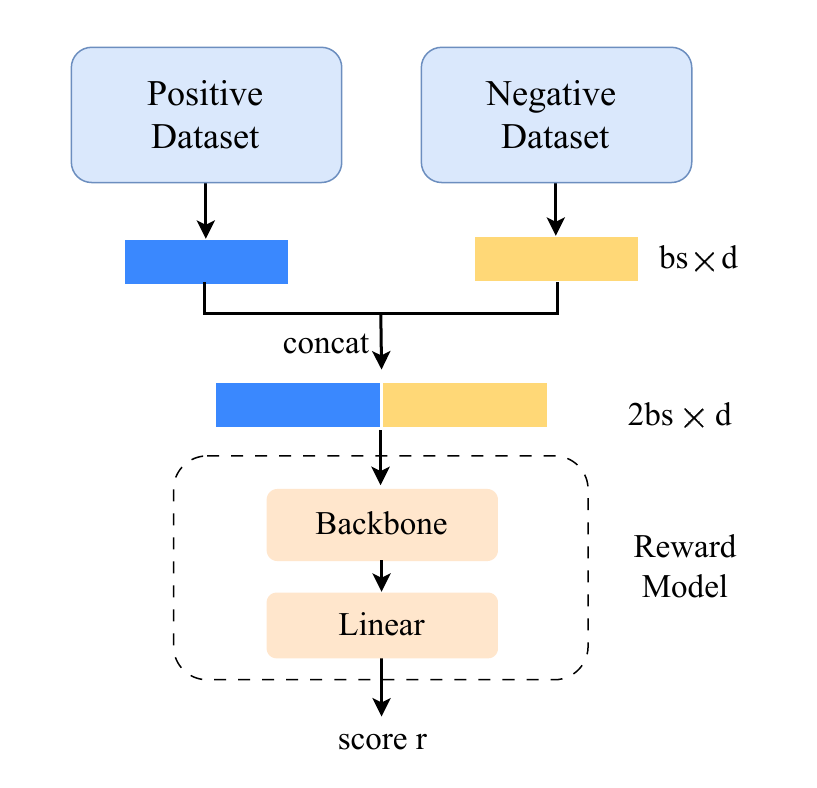}
	\label{fig:2}
	\caption{Structure of reward model in stage one. The reward model is composed of the backbone and linear layer mapping features to a scalar $r$. Positive and negative sample pairs are taken as input. The model is trained to obtain higher $r$ for positive samples than negative samples.}
\end{figure}

\begin{figure*}[!h]
	\centering
	\includegraphics[scale=0.6]{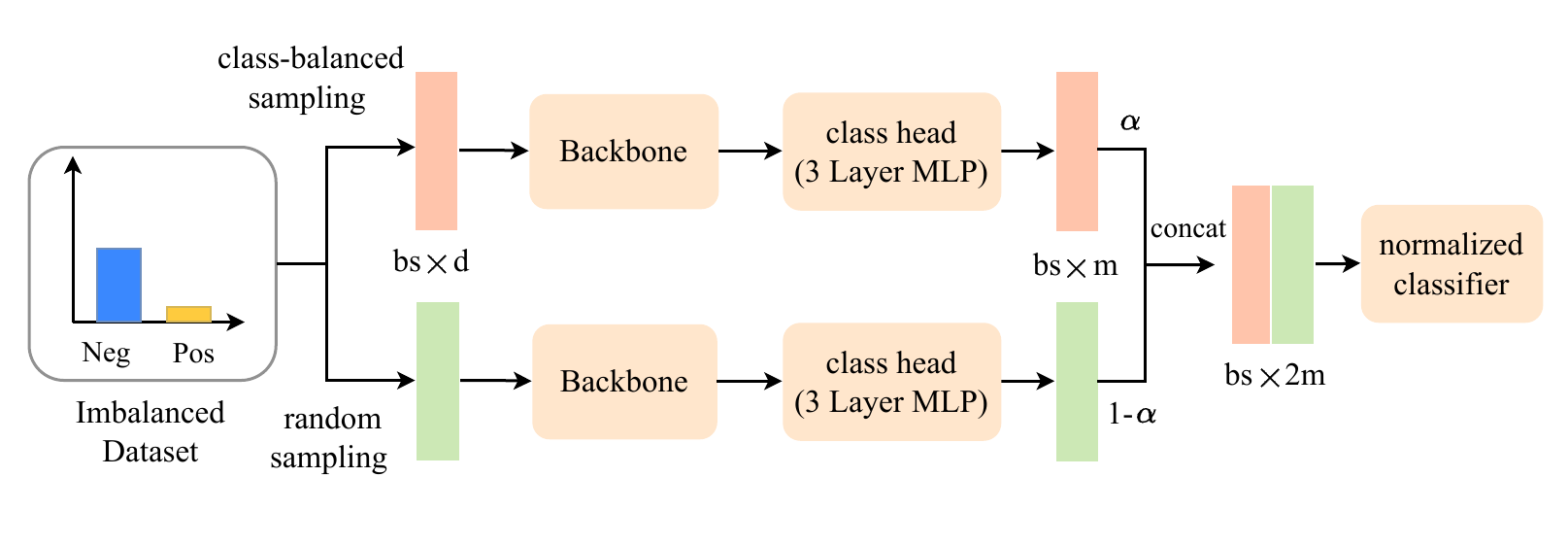}
	\label{fig:3}
	\caption{Structure of enhanced BBN in stage two. The upper branch focus on the representation learning of positive samples with class-balanced sampling and focal loss, while the lower branch focus on the representation learning of negative samples by uniform sampling and cross-entropy loss. Then features is mixed with adaptive parameter $a$. Finally a normalized classifier is adopted to mitigate the model's preference for negative samples. }
\end{figure*}

\section{METHODS}

In this section, we propose a multi-stage training framework to progressively densify the dataset and alleviate the degree of imbalance, which consists of three stages. 
In the first stage, we utilize unsupervised learning model to pre-identify those non-critical negative samples. In this way, we can effectively reduce the imbalance ratio.
In the second stage, labeled data is leveraged to train a classification model capable of further distinguishing between positive samples and hard negative samples. 
Finally, in the third stage, we employ dense reinforcement learning to improve the accuracy of criticality predicted by the classification model. 

\subsection{Stage One: Ranking Loss Based Unsupervised Learning} 

When confronted with a scarcity of positive samples, effectively capturing their distribution becomes a challenging task. In such scenarios, some approaches exploit anomaly detection to learn the distribution of negative samples and identify samples that deviate from this distribution as potential positive instances \cite{xu2022constructing}. In this work, we leverage self-supervised learning to train a reward model. This model aims to increase the distance between positive and negative samples in the feature space. By doing so, we can identify samples that exhibit clear non-critical characteristics. Free from the influence of \enquote{label bias}\cite{liu2021self}, the reward model can learn more robust features. 

Fig. 2 illustrates the architecture of reward model $r_{\theta}$. During the training phase, the model takes pairs of positive and negative sample $(\bm{x}_{pi},\bm{x}_{ni})$ as input. The objective of this reward model is to assign a scalar $r$ to each input sample, with the purpose that positive samples get higher scores than negative ones.
The loss function is defined as follows:
\begin{equation}
	\mathcal{L}(\theta) = -\mathbb{E} _{\bm{X}_p\sim \mathcal{P}, \bm{X}_n\sim \mathcal{N}} \left[\log \sigma (r_{\theta}(\bm{X}_p) - r_{\theta}(\bm{X}_n))\right], 
\end{equation}
where $\bm{X}_p$ represents the positive samples and $\bm{X}_n$ denotes the negative samples. $\sigma$ denotes the sigmoid function. This model essentially learns a mapping relationship between the feature space and the scalar space. By selecting an appropriate threshold $\epsilon$, we can effectively identify the \enquote{easy} negative samples or certainly non-critical samples, while retaining the positive samples. This means that a sample $x$ is considered positive if $r_{\theta}(x)>\epsilon$ and negative otherwise.  Since our aim is to improve precision rate without compromising recall rate, the threshold $\epsilon$ must guarantee that most positive samples are retained.  

In addition, although performance of the reward model maybe affected by the extent of class overlap, employing unsupervised models allows us to substantially decrease the number of negative samples, thereby reducing the imbalance ratio of dataset.

\subsection{Stage Two: Enhanced BBN based Supervised Learning}
In the second stage of our framework, we employ supervised learning to train a classification model. It is worth noting that in this stage, our positive dataset $\mathcal{P}$ includes all positive samples utilized in the first stage. In contrast, our negative dataset $\mathcal{N}^-$ exclusively comprises false positive samples indentified in the first stage. We adopt a more complex model as the difficulty of samples increases compared to the first stage.

Inbalanced test dataset poses greater challenges for learned representation, and common rebalancing strategies such as re-sampling and re-weighting are often detrimental to feature learning. Intuitively, random sampling might lead to better representations. To address this issue, we develop an enhanced BBN method to improve the representation learning.  As depicted in Fig. 3, in the upper branch, we utilize class-balanced sampling to focus on learning representations of positive samples. Simultaneously, in the lower branch, we employ random sampling to extract features from negative samples. The features from both branches are then combined as follows:
\begin{equation}
	 \bm{z} = \alpha \bm{W}_a^\mathrm{T} \bm{f}_a + (1-\alpha)\bm{W}_b^\mathrm{T} \bm{f}_b,  
\end{equation}
 where $\bm{z}$ represents the mixed feature representation, $\alpha$ is a hyper parameter controlling the mixing ratio, $\bm{W}_a$ and $\bm{W}_b$ are the weight matrices for the respective branches, and $\bm{f}_a$ and $\bm{f}_b$ are the extracted features. Empirical observations suggest that the positive class tends to have larger classifier norms\cite{thrampoulidis2022imbalance}. To mitigate the classifier's bias toward the majority class, we adopt a normalized linear classifier. This normalization ensures that the classifier's weights are scaled to have unit norm, effectively balancing the influence of different classes. The normalized weights are computed as
 \begin{equation}
 	 \hat{\bm{w}_i} = \frac{\bm{w}_i}{||\bm{w}_i||^\mathrm{T}}, 
 \end{equation}
where $ \hat{\bm{w}_i} $ represents the normalized weight and $\bm{w}_i$ is the original weight. The loss function is given by:
 \begin{equation}
	 \mathcal{L} = \alpha \mathcal{L}_a(\bm{p},\bm{y}_a) + (1-\alpha) \mathcal{L}_b(\bm{p},\bm{y}_b), 
\end{equation}
 where $\bm{p}$ represents logits of classification model, and $\bm{y}_a$ and $\bm{y}_b$ are the corresponding labels for the two branches. $\mathcal{L}_a$ is the focal loss to further enhance feature learning of positive samples, and $\mathcal{L}_b$ is the cross-entropy loss. Using the enhanced BBN, we can further categorize samples that failed to be distinguished in the first stage. As the imbalance ratio decreases, the difficulty of training in this stage can be mitigated.

 \subsection{Stage Three: Dense Reinforcement Learning}
 
Since our primary objective is to achieve precise criticality predictions, our focus in this stage shifts from continuously categorizing those hard or even non-separable samples to fine-tuning the classification model to further improve the accuracy of criticality assessments. Specifically, we employ dense reinforcement learning\cite{feng2023dense} to fine-tune the last layers of classification model in stage two, including the class head and classifier.

In reinforcement learning, the value of each state-action pair, denoted as $Q(\bm{s}, \bm{a})$, represents the expected return obtained by executing action $a$ in state $s$.  
To learn the Q-value, we utilize a dense deep Q-learning (DQN) approach, where the classifier serves as the Q-network and the predicted criticality represents the Q-value
 \begin{equation}
	Q(\bm{s}, \bm{a}) \triangleq \mathbb{P}(A|\bm{s},\bm{a}) = \mathbb{P}(A|\bm{\bm{X}}).
\end{equation}
 In practical applications, we introduce historical states to enrich the input information for improved classification accuracy. Consequently, the input $\bm{X}$ in previous stages contains not only $\bm{s}$ and $\bm{a}$ at current step but also previous states. For simplicity,  we represent the input $\bm{X}$ as $(\bm{s},\bm{a})$ in the rest of the paper. We adopt an offline training approach for efficiency. 
 
 The term \enquote{dense} refers to our use of critical episodes only for training Q-network. An episode, denoted as $(\bm{s}_{i1}, \bm{s}_{i2},...,\bm{s}_{it})$, is considered critical if a critical event $A$ occurs at $\bm{s}_{it}$.  According to our definition of train dataset $\mathcal{D}$, the state $\bm{s}_{it}$ is a positive sample and $\bm{s}_{ij}$ for $j=1,...,t-1,$ are negative samples. Additionally, states belonging to non-critical episodes are also considered as negative samples. This allows us to focus on the most relevant and informative data for fine-tuning the classifier. Our reward function is defined as follows
\begin{equation}
	 r(\bm{s}) = \mathbb{I}_A(\bm{s}),
\end{equation}
 where $r=1$ if at state $s$ critical event happen and 0 otherwise. This reward function encourages the Q-network to assign higher Q-values to states that lead to critical outcomes. The Q-network is updated according to the following loss function 
 \begin{equation}
 	 \mathcal{L}(\bm{\theta}_k)=\sum_{(\bm{s},\bm{a},r,\bm{s}')\in \mathcal{D}} (y-Q(\bm{s},\bm{a};\bm{\theta}_k))^2\mathbb{I}_{\bm{s}\in \mathbb{S}_c},
 \end{equation}
 where $y=r+\gamma \max_{\bm{a}'} Q(\bm{s}',\bm{a}';\bm{\theta}_{k-1})$ and $\mathbb{S}_c$ denotes the set of states belonging to critical episodes.

By employing dense DQN, we can optimize the criticality predicted by classifier. Due to the large of number of negative samples, the gradients are dominated by negative samples, leading to the classifier biased toward the negative class\cite{molahasani2023continual}. With dense DQN, we are able to fine-tune classification model with more balanced gradient.
The derivative of the loss function is
\begin{equation}
	\frac{\nabla \mathcal{L}(\bm{\theta}_k)}{\nabla \bm{\theta}_k} = -2\sum_{(\bm{s},\bm{a},r,\bm{s}')\in \mathcal{D}}(y-Q(\bm{s},\bm{a};\bm{\theta}_k))\frac{\nabla  Q(\bm{s},\bm{a};\bm{\theta}_k)}{\nabla \bm{\theta}_k}\mathbb{I}_{\bm{s}\in \mathbb{S}_c}
\end{equation}

Then the loss function $\mathcal{L}(\bm{\theta}_k)$ can be reformulated as the sum of the positive class loss function $\mathcal{L}_p(\bm{\theta}_k)$ and the negative class loss function $\mathcal{L}_n(\bm{\theta}_k)$, we have
\begin{equation}
	\begin{aligned}
		\frac{\nabla \mathcal{L}(\bm{\theta}_k)}{\nabla \bm{\theta}_k} &= 	\frac{\nabla \mathcal{L}_p(\bm{\theta}_k)}{\nabla \bm{\theta}_k} + \frac{\nabla \mathcal{L}_n(\bm{\theta}_k)}{\nabla \bm{\theta}_k} \\
	&= 2\sum_{i=1}^{N_p}(f_{\bm{\theta}_k}(\bm{s}_i,\bm{a}_i)-1)\frac{\nabla f_{\bm{\theta}_k}}{\nabla \bm{\theta}_k}+\\
	&2\sum_{j=1}^{N_n'}(f_{\bm{\theta}_k}(\bm{s}_j,\bm{a}_j) - \gamma \max_{\bm{a}_j'} f_{\theta_{k}}(\bm{s}_j',\bm{a}_j'))\frac{\nabla f_{\bm{\theta}_k}}{\nabla \bm{\theta}_k} 
	\end{aligned}
\end{equation}
where $f_{\bm{\theta}_k}$ is the classification model. We can observed that $N_p$ equals $|\mathcal{P}|$ and the number of negative samples $N_n'\ll |\mathcal{N}|$ due to the dense method. Therefore, in comparison to the second stage, we further alleviate the impact of negative samples on the loss function. This helps in reducing the classifier's preference for negative samples and improving the precision of criticality.

\section{RESULTS}

In this section, we applied our approach to two well-known cases: Bipedal Walker and Lunar Lander, both of which are tasks under the Gym environment. Next, we provide a brief overview of each case and present the corresponding ROC curves and Precision-Recall curves. Eventually, ablation studies are performed to verify the effectiveness of each components of our method.
\begin{figure}[!h]
	\centering
	\begin{minipage}{1\linewidth}	% linewidth就是栏宽
		\subfigure[Case of lunar lander]{
			\label{fig:g}
			\includegraphics[width=0.48\linewidth,height=1in]{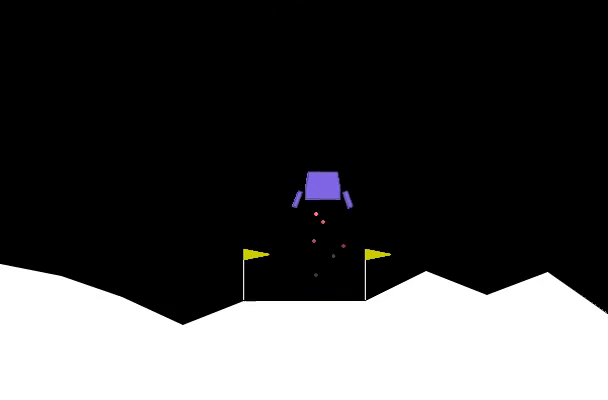}	
		}\noindent
		\subfigure[Case of bipedal walker]{
			\label{fig:h}
			\includegraphics[width=0.48\linewidth,height=1in]{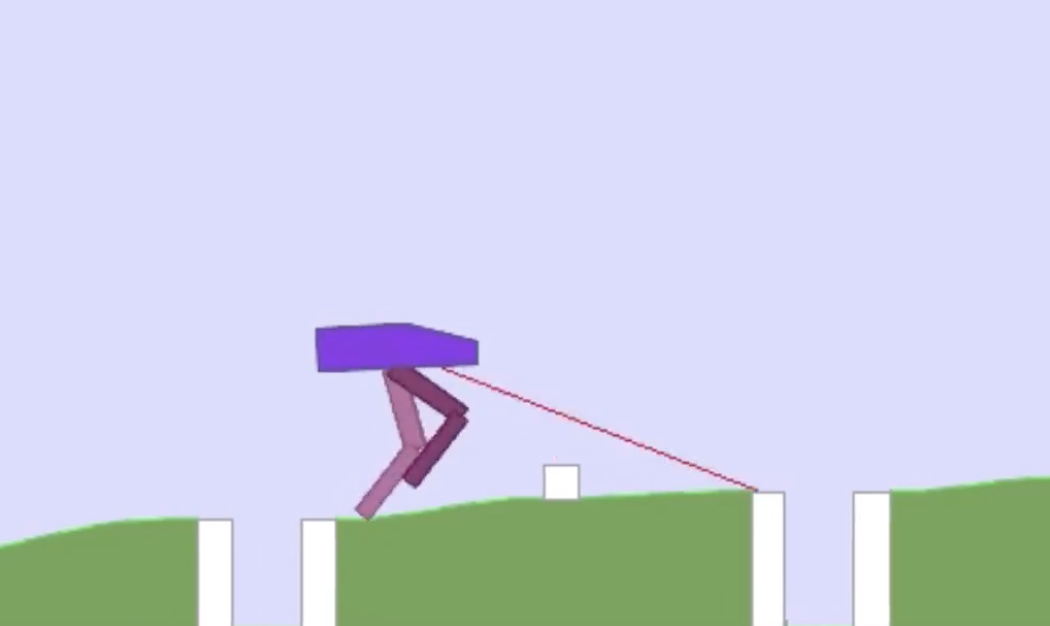}	}
	\end{minipage}
	\vspace{-0.18in}	% 调整大标题和图片之间的距离，单位有cm in pt
	\caption{Overview of cases}
	\vspace{-0.2in}		% 调整正文部分和标题（图片之间的距离）
	\label{fig:5}
\end{figure}

 \begin{figure*}[!h]
	\centering
	\begin{minipage}{1\linewidth}	% linewidth就是栏宽
		\subfigure[Results of lunar lander case]{
			\label{fig:e}
			\includegraphics[width=0.49\linewidth,height=2in]{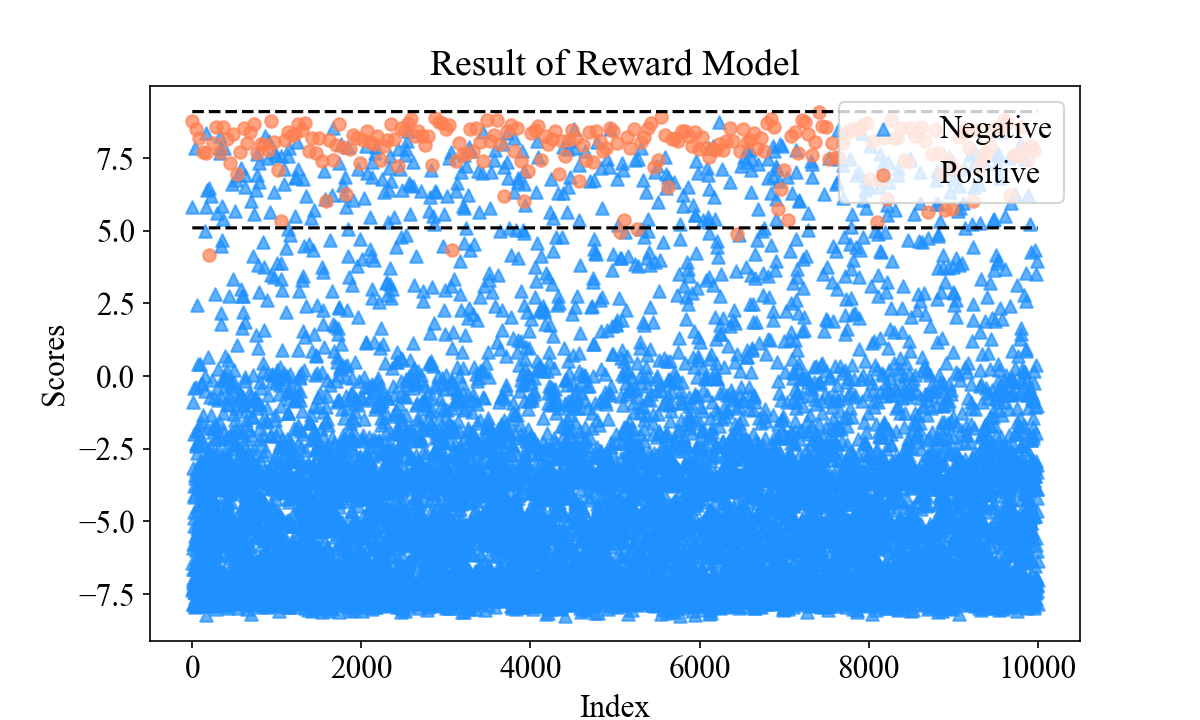}	
		}\noindent
		\subfigure[Results of bipedal walker case]{
			\label{fig:f}
			\includegraphics[width=0.49\linewidth,height=2in]{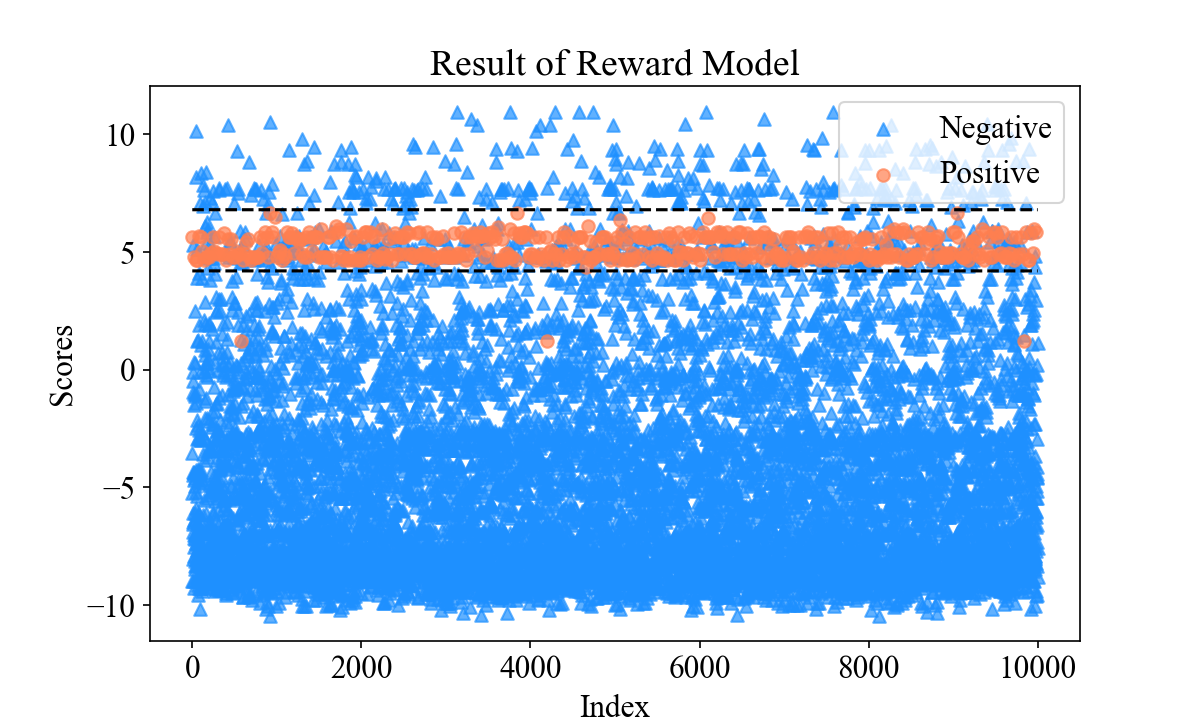}
		}
	\end{minipage}
	\vspace{-0.18in}	% 调整大标题和图片之间的距离，单位有cm in pt
	\caption{Outputs of reward model}
	\vspace{-0.2in}		% 调整正文部分和标题（图片之间的距离）
	\label{fig:7}
\end{figure*}

\begin{figure*}[!h]
	\centering
	\begin{minipage}{1\linewidth}	% linewidth就是栏宽
		\subfigure[ROC curve of lunar lander case]{
			\label{fig:a}
			\includegraphics[width=0.49\linewidth,height=2in]{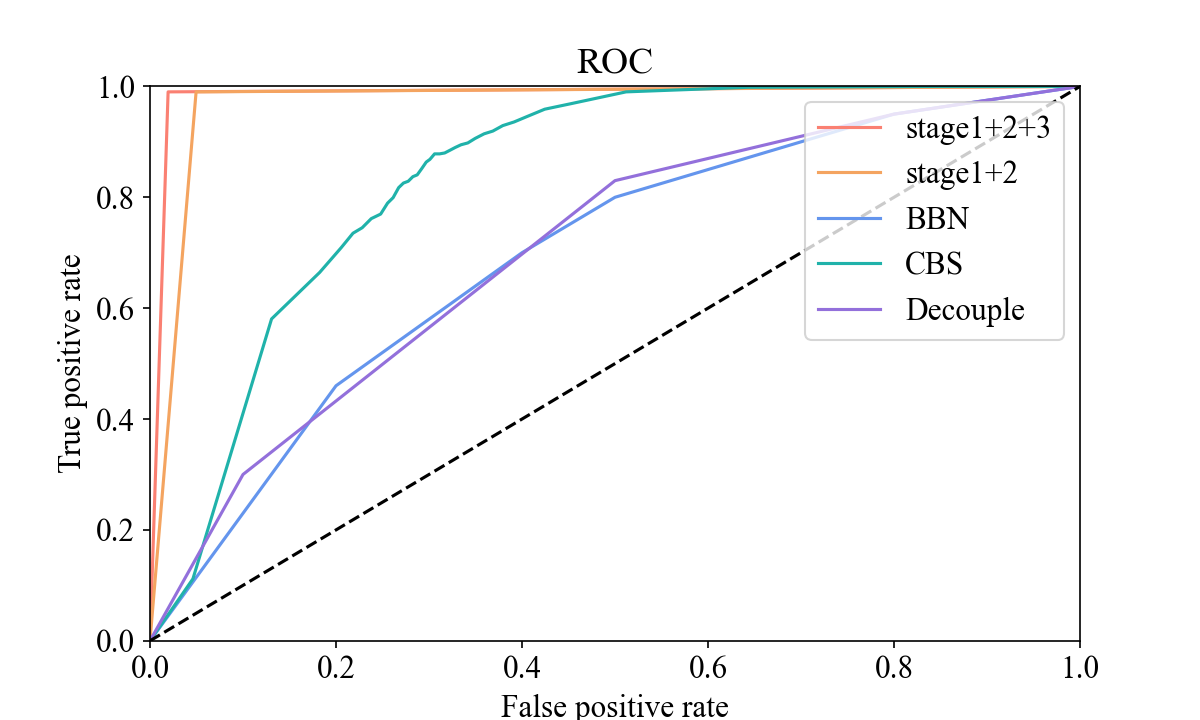}	
		}\noindent
		\subfigure[PR curve of lunar lander case]{
			\label{fig:b}
			\includegraphics[width=0.49\linewidth,height=2in]{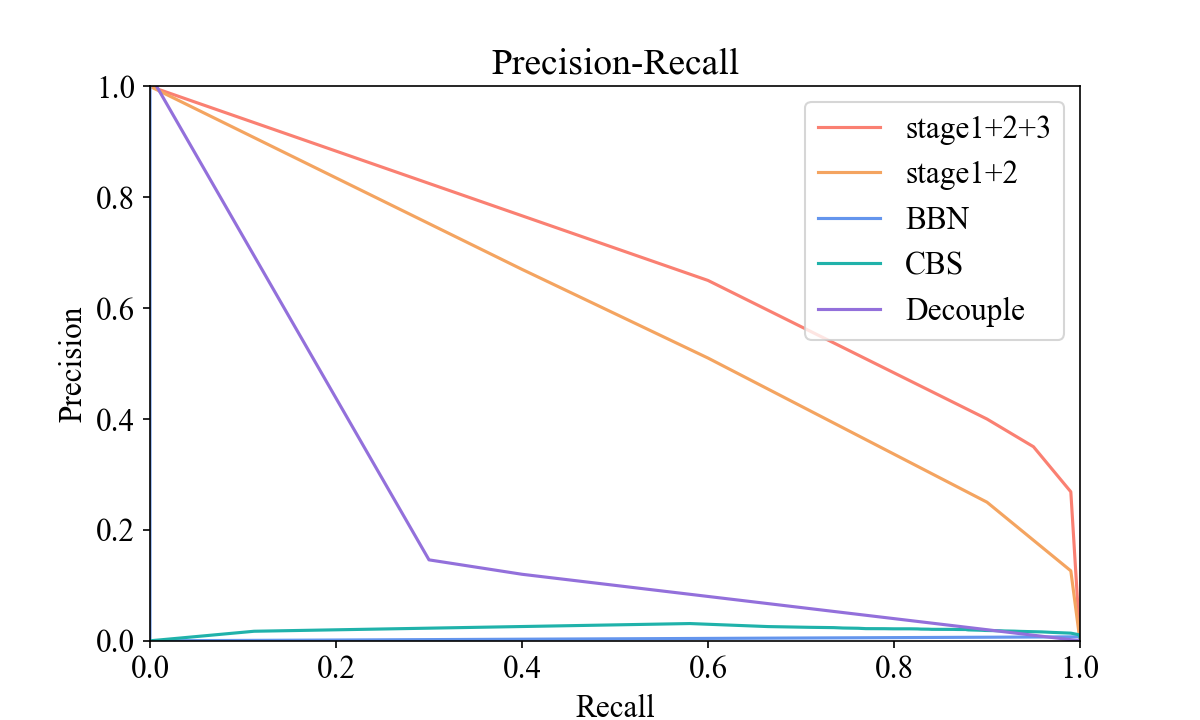}
		}
	\end{minipage}
	\vskip -0.3cm % 用于调整两个minipage之间的垂直间距
	\begin{minipage}{1\linewidth }
		\subfigure[ROC curve of bipedal walker case]{
			\label{fig:c}
			\includegraphics[width=0.49\linewidth,height=2in]{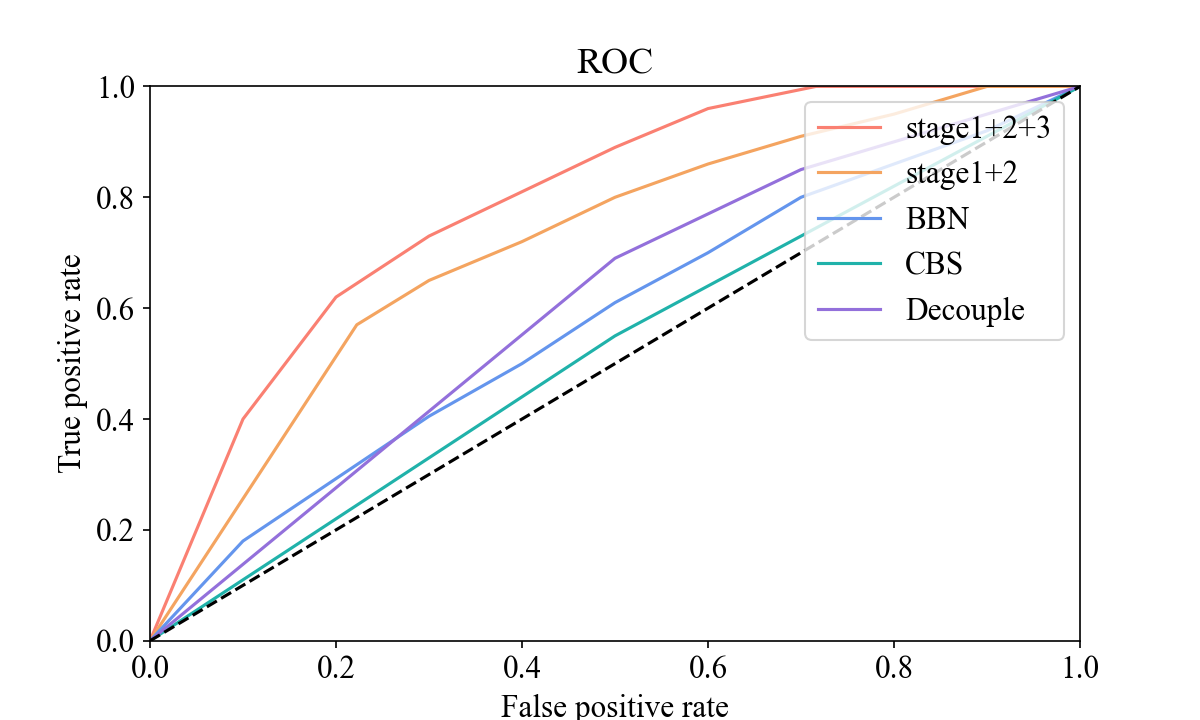}
		}\noindent
		\subfigure[PR curve of bipedal walker case ]{
			\label{fig:d}
			\includegraphics[width=0.49\linewidth,height=2in]{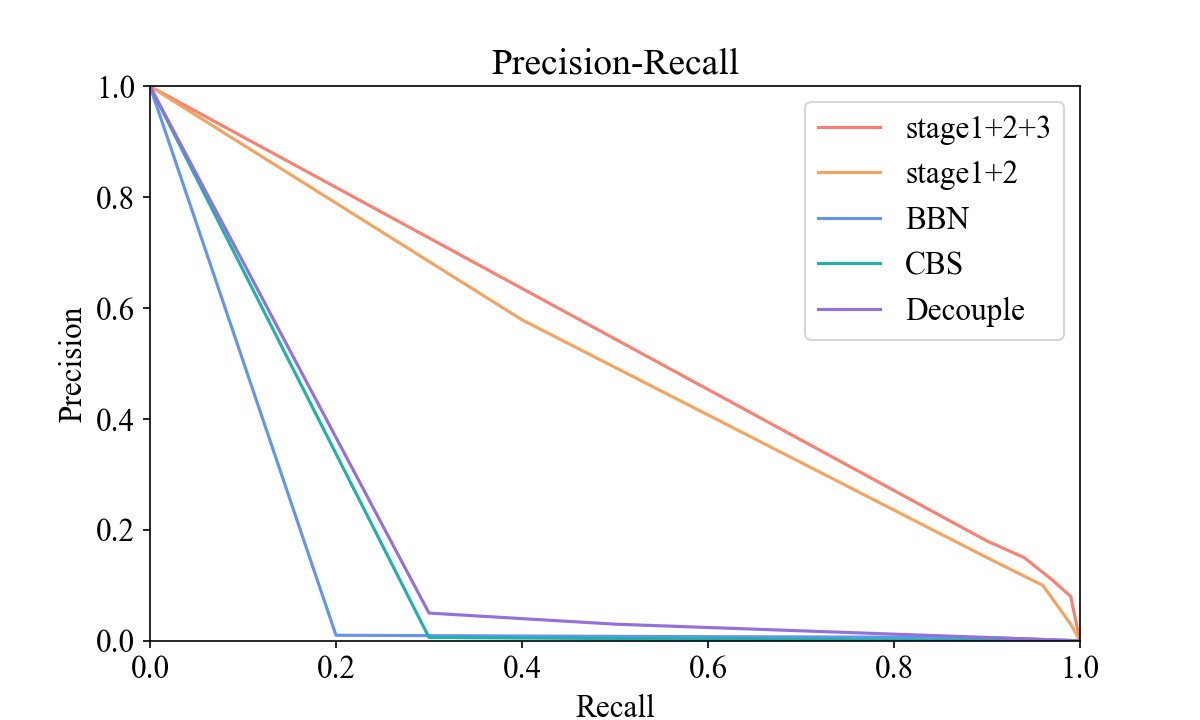}
		}
	\end{minipage}
	\vspace{-0.18in}	% 调整大标题和图片之间的距离，单位有cm in pt
	\caption{ROC curves and Precision-Recall curves of supervised classification model and dense DQN fine-tuned model}
	\vspace{-0.2in}		% 调整正文部分和标题（图片之间的距离）
	\label{fig:4}
\end{figure*}
\subsection{Overview of Cases}

In the lunar lander case, as shown in Fig. 4(a), a rocket is optimized to achieve a safe touchdown on the landing pad, avoiding any potential crashes. The criticality event $A$ occurs when the lander gets in contact with the moon. A well-trained rocket using PPO algorithm was utilized to collect the training dataset. Considering information related to the rocket's state and atmospheric conditions, the input feature $X_k$ is defined as 
\begin{equation}
	\bm{X}_k = (\bm{s}_{k-10},.., \bm{s}_{k-1}, \bm{s}_k),
\end{equation}
where $\bm{s}_k \in \mathbb{R}^{11}$ represents the state at step $k$, and $\bm{s}_{k-10},...,\bm{s}_{k-1}$ represent historical states.  The state $\bm{s}_k$ consists of both the rocket's state $\bm{s}_{kr}$  and the environment's state $\bm{s}_{ke}$, where $\bm{s}_{kr}$ includes the rocket's coordinates, linear velocities, angles, angular velocity, along with two boolean values indicating whether each leg is in contact with the ground. Additionally, $\bm{s}_{kr}$ includes the force of linear wind and rotational wind. If critical event occurs at terrain $\bm{s}_{k}$, the sample $\bm{x}_k$ is labeled as a positive sample $(y_i=1)$. The original train dataset exhibits an imbalanced ratio of $1.26\times 10^4$.

In the bipedal walker case, as shown in Fig. 4(b), a four joint walker must navigate through a series of obstacles, including  stumps and pitfalls, to reach its destination. The criticality event $A$ is defined as the hull of walker contact with the ground without achieving a predefined distance. Utilizing the POET\cite{wang2019paired} algorithm, a well-trained walker was developed. The input feature $\bm{X}_k$ is

\begin{equation}
	\bm{X}_k = (\bm{s}_{k0}, \bm{s}_{k1}, \bm{s}_{k2},..., \bm{s}_{k9} ),
\end{equation}
where $\bm{s}_{k0} \in \mathbb{R}^{25}$ represents the walker's state, including hull angle speed, angular velocity, horizontal speed, vertical speed, position of joints and joints angular speed, legs contact with ground, and lidar measurements. $\bm{s}_{ki} \in \mathcal{R}^8$ represents terrain within the detection range of lidar. If the critical event occurs at terrain $\bm{s}_{k9}$, the sample $\bm{X}_k$ is considered as positive sample with $y_k=1$. The imbalanced ratio of train dataset approaches $1.56\times 10^4$.

\subsection{Results}

For the Lunar Lander case, in the first stage, a 6-layer transformer serves as the backbone of our reward model.  As depicted in Fig. \ref{fig:e}, orange points represent results of positive samples, while blue points represent results of negative ones. The dashed lines indicate the thresholds we select. We manage to delete 95.02\% negative samples and identify 99.27\% positive samples. The backbone used in the second stage is identical to that of reward model. Using both the reward model and classification model, we can achieve improved identification rates, 98.41\% for negative samples and 99.13\% for positive samples, as well as the AUC of 0.9612. Finally, dense DQN leads to a further improvement: 99.06\% for negative samples and 99.18\% for positive samples, resulting in AUC increase to 0.9853. 

For the bipedal walker case, in the first stage, a 6-layer transformer serves as backbone of our reward model.  As shown in Fig. \ref{fig:f}, by selecting a thresholds of 4.2 and 6.8, we can eliminate 79.35\% negative samples and identify 87.60\% positive samples. Then, benefiting from classification model, we can identify 88.72\% negative samples and 90.12\% positive samples and the AUC is 0.6752. By employing dense DQN, we can identify 91.66\% negative samples and 93.09\% positive samples and the AUC is 0.7163.

As illustrated in Fig. \ref{fig:4}, we compared our method with traditional approaches including original BBN\cite{zhou2020bbn}, class-balanced sampling (CBS)\cite{cui2019class} and decoupling method\cite{kang2019decoupling}. It is evident that those classical methods mentioned above may not be effective when dealing with extremely imbalance test dataset. Additionally, removing certain non-critical samples with unsupervised learning stage proved beneficial in improving identification rates. Furthermore, it can also be demonstrated that performance gains further improvement with dense DQN. 

%It is worth noting that when dealing with extremely imbalanced training datasets, typical approaches mentioned %above may not be effective if the test dataset also exhibits the same imbalance. 
%%%
%\begin{table}[!h]
%\caption{An Example of a Table}
%\label{table_example}
%\begin{center}
%\begin{tabular}{|c||c|}
%\hline
%One & Two\\
%\hline
%Three & Four\\
%\hline
%\end{tabular}
%\end{center}
%\end{table}
%%%

\section{CONCLUSIONS}
In conclusion, our study focuses on developing a criticality prediction model for intelligent systems that achieved both high precision and recall rates. To address the challenge of curse of rarity, we propose a multi-stage learning framework, which gradually densifies the dataset and alleviates the degree of imbalance through a combination of unsupervised learning and reinforcement learning methods. Our research provides valuable insights into the criticality prediction of intelligent systems, thus facilitating their development and real-world deployment. In the future, we will delve deeper into the theoretical underpinnings of our proposed method and determine the imbalance ratio for which it is applicable.


%%%%%%%%%%%%%%%%%%%%%%%%%%%%%%%%%%%%%%%%%%%%%%%%%%%%%%%%%%%%%%%%%%%%%%%%%%%%%%%%

%%%%%%%%%%%%%%%%%%%%%%%%%%%%%%%%%%%%%%%%%%%%%%%%%%%%%%%%%%%%%%%%%%%%%%%%%%%%%%%%

%%%%%%%%%%%%%%%%%%%%%%%%%%%%%%%%%%%%%%%%%%%%%%%%%%%%%%%%%%%%%%%%%%%%%%%%%%%%%%%%

%\section*{APPENDIX}

%\section*{ACKNOWLEDGMENT}

%%%%%%%%%%%%%%%%%%%%%%%%%%%%%%%%%%%%%%%%%%%%%%%%%%%%%%%%%%%%%%%%%%%%%%%%%%%%%%%%
\bibliographystyle{IEEEtran}
\bibliography{reference.bib}

%\begin{thebibliography}{1}
%\bibitem{1} 
%\end{thebibliography}
\end{document}